\pdfoutput=1

\documentclass[11pt]{article}

\usepackage[]{EMNLP2022}

\usepackage{times}
\usepackage{latexsym}

\usepackage[T1]{fontenc}

\usepackage[utf8]{inputenc}

\usepackage{microtype}

\usepackage{inconsolata}

\usepackage{xspace}
\usepackage{graphicx}
\usepackage[inline]{enumitem}  
\usepackage[group-separator={,}]{siunitx}  
\usepackage{booktabs}
\usepackage{multirow}
\usepackage{floatrow}
\usepackage{amsmath}

\title{Entity Disambiguation with Entity Definitions}

\author{Luigi Procopio$^1$ \qquad Simone Conia$^1$ \qquad Edoardo Barba$^1$ \qquad Roberto Navigli$^2$\\
         Sapienza NLP Group\\
         Sapienza University of Rome\\
         $^1$\texttt{\{lastname\}@di.uniroma1.it}\\
         $^2$\texttt{navigli@diag.uniroma1.it}}

\begin{document}
\maketitle
\begin{abstract}

Local models have recently attained astounding performances in Entity Disambiguation (ED), with generative and extractive formulations being the most promising research directions. However, previous works limited their studies to using, as the textual representation of each candidate, only its Wikipedia title. Although certainly effective, this strategy presents a few critical issues, especially when titles are not sufficiently informative or distinguishable from one another. In this paper, we address this limitation and investigate to what extent more expressive textual representations can mitigate it. We thoroughly evaluate our approach against standard benchmarks in ED and find extractive formulations to be particularly well-suited to these representations: we report a new state of the art on \num{2} out of \num{6} benchmarks we consider and strongly improve the generalization capability over unseen patterns. We release our code, data and model checkpoints at \url{https://github.com/SapienzaNLP/extend}.

\end{abstract}

\section{Introduction}

Being able to pair a mention in a given text with its correct entity out of a set of candidates is a crucial problem in Natural Language Processing (NLP), referred to as Entity Disambiguation \cite[ED]{bunescu-pasca-2006-using}. Indeed, since ED enables the identification of the actors involved in human language, it is often considered a necessary building block for a wide range of downstream applications, including Information Extraction \cite{ji-grishman-2011-knowledge,guo-etal-2013-link}, Question Answering \cite{yin-etal-2016-simple} and Semantic Parsing \cite{bevilacqua-etal-2021-one,procopio-etal-2021-sgl}. ED generally occurs as the last step in an Entity Linking pipeline \cite{broscheit-2019-investigating}, preceded by Mention Detection and Candidate Generation, and its approaches have been traditionally divided into two groups, depending on whether co-occurring mentions are disambiguated independently (\textit{local methods}; \citet{shahbazi-etal-2019-entity,wu-etal-2020-scalable,tedeschi-etal-2021-named-entity}) or not (\textit{global methods}; \citet{hoffart-etal-2011-robust,moro-etal-2014-entity,yamada-etal-2016-joint,yang-etal-2018-collective}).

Despite the limiting operational hypothesis of independence between co-occurring mentions, local methods have nowadays achieved performances that are either on par or above those attained by their global counterparts, mainly thanks to the advent of large pre-trained language models. In particular, among these methods, generative \cite{decao2020autoregressive} and extractive \cite{barba-etal-2022-extend} formulations are arguably the most promising directions, having resulted in large performance improvements across multiple benchmarks. Regardless of their modeling differences, the key idea behind these methods is to part away from the previous classification-based approaches and, instead, adopt formulations that better leverage the original pre-training of the underlying language models. On the one hand, generative formulations tackle ED as a text generation problem and train neural architectures to auto-regressively generate, given a mention and its context, a textual representation of the correct entity. On the other hand, extractive approaches frame ED as extractive question answering: they first concatenate a textual representation of each entity candidate to the original input and then train a model to extract the span corresponding to the correct entity.

Although having admittedly attained great improvements, both in- and out-of-domain, to the best of our knowledge, previous works on both these formulations have limited their studies to a single type of textual representation for entities, that is, their title in Wikipedia. However, this strategy presents a number of issues \cite{barba-etal-2022-extend} and, in particular, often results in representations that are either insufficiently informative or even virtually indistinguishable between one another. In contrast to this trend, we address this limitation and explore the effect of more expressive textual representation on state-of-the-art local methods. To this end, we propose to complement Wikipedia titles with their description in Wikidata so that, for instance, the candidates for \textit{Ronaldo} in \textit{\underline{Ronaldo} scored two goals for Portugal} would be \textit{Cristiano Ronaldo: Portoguese association football player} and \textit{Ronaldo: Brazilian association football player}, rather than the less informative \textit{Cristiano Ronaldo} and \textit{Ronaldo}. We test our novel representations on generative and extractive formulations, and evaluate against standard benchmarks in ED, both in and out of domain, reporting statistically significant improvements for the latter group.

\section{Method}

We now formally introduce ED and the textual representation strategy we put forward. Then, we describe the two formulations with which we implement and test our proposal.

\paragraph{ED with Entity Definitions} Given a mention $m$ occurring in a context $c_m$, Entity Disambiguation is formally defined as the task of identifying, out of a set of candidates $e_1, \dots, e_n$, the correct entity $e^{*}$ that $m$ refers to. In generative and extractive formulations, each candidate $e$ is additionally associated with a text representation $\hat{e}$, which is a string describing its meaning. Whereas previous works have considered the title that $e$ had in Wikipedia as $\hat{e}$, here we focus on more expressive alternatives and leverage Wikidata to achieve this objective. In particular, we first retrieve the Wikidata description of $e$. Then, we define as the new representation of $e$ the colon-separated concatenation of its Wikipedia title and its Wikidata description, e.g., \textit{Ronaldo: Brazilian association football player}.

\paragraph{Generative Modeling}  In our first formulation, we follow \citet{decao2020autoregressive} and frame ED as a text generation problem. Starting from a mention $m$ and its context $c_m$, we first wrap the location of $m$ in $c_m$ between two special symbols, namely \textit{<s>} and \textit{</s>}; we denote this modified sequence by $\tilde{c}_m$. Then, we train a sequence-to-sequence model to generate the textual sequence $\hat{e}^{*}$ of the correct entity $e^{*}$ by learning the following probability:
\begin{align*}
    p(\hat{e}^{*} | \tilde{c}_m) = \prod_{j=1}^{|\hat{e}^{*}|} p(\hat{e}^{*}_j | \hat{e}^{*}_{1: j - 1}, \tilde{c}_m)
\end{align*}
where $\hat{e}^{*}_j$ denotes the $j$-th token of $\hat{e}^{*}$ and $\hat{e}^{*}_{0}$ is a special start symbol. The purpose of \textit{<s>} and \textit{</s>} is to signal the model that $m$ is the token we are interested in disambiguating. As in the reference work, we use BART \cite{lewis-etal-2020-bart} as our sequence-to-sequence architecture for our experiments and, most importantly, adopt constraint decoding on the candidate set at inference time. Indeed, applying standard decoding methods such as beam search might result in outputs that do not match any of the original candidates; thus, to obtain only valid sequences, at each generation step, we constrain the set of tokens that can be generated according to a prefix tree \cite{DBLP:books/daglib/0023376} built over the candidate set.

\paragraph{Extractive Modeling} Additionally, we also consider the formulation recently presented by \citet{barba-etal-2022-extend} that frames ED as extractive question answering. Here, $\tilde{c}_m$, defined analogously to the previous paragraph, represents the query, whereas the context is built by concatenating a textual representation of each candidate $e_1, \dots, e_n$. A model is then trained to extract the text span that corresponds to $e^{*}$. Following the efficiency reasoning of the authors, we use as our underlying model the Longformer \cite{beltagy2020longformer}, whose linear attention better scales to this type of long-input formulations. Compared to the above generative method, the benefits of this approach lie in
\begin{enumerate*}[label=\roman*)]
    \item dropping the need for a potentially slow auto-regressive decoding process and
    \item enabling full joint contextualization both between context and candidates and across candidates themselves.
\end{enumerate*}

\section{Experiments and Results}

\begin{table}[t]

    %
    %
    %
    %
    %
    %
    %
    %

    \centering
    \resizebox{1.0\linewidth}{!}{
    
        \begin{tabular}{clrrr}
        \toprule
        & Dataset & Instances & Candidates & Failures \\
        
        \toprule
        
        \parbox[t]{1mm}{\multirow{3}{*}{\rotatebox[origin=c]{90}{\textit{AIDA}}}}
        & Train         & \num{18448} & \num{905916}$^{\, / \, \num{79561}}$ & \num{5038}$^{\, / \, \num{682}}$ \\
        & Validation    & \num{4791} & \num{236193}$^{\, / \, \num{43339}}$ & \num{1360}$^{\, / \, \num{296}}$ \\
        & Test          & \num{4485} & \num{231595}$^{\, / \, \num{46660}}$ & \num{1395}$^{\, / \, \num{323}}$ \\
        
        \midrule
        
        \parbox[t]{1mm}{\multirow{5}{*}{\rotatebox[origin=c]{90}{\textit{OOD}}}}
        & MSNBC         & \num{656} & \num{17895}$^{\, / \, ~~~~\num{8336}}$ & \num{149}$^{\, / \, ~~~~\num{72}}$ \\
        & AQUAINT       & \num{727} & \num{23917}$^{\, / \, ~~\num{16948}}$ & \num{142}$^{\, / \, ~~\num{121}}$ \\
        & ACE2004       & \num{257} & \num{12292}$^{\, / \, ~~~~\num{8045}}$ & \num{66}$^{\, / \, ~~~~\num{50}}$ \\
        & CWEB          & \num{11154} & \num{462423}$^{\, / \, \num{119781}}$ & \num{3642}$^{\, / \, \num{1265}}$ \\
        & WIKI          & \num{6821} & \num{222870}$^{\, / \, \num{105440}}$ & \num{1216}$^{\, / \, ~~\num{719}}$ \\
        
        \bottomrule
        \end{tabular}
        
    }
    
    \caption{Number of instances, candidates and failures to map a Wikipedia title to its Wikidata definition in the AIDA-CoNLL (top) and out-of-domain (bottom) datasets. For candidates and failures, we report both their total (base) and unique (exponent) number.}
    \label{tab:datasets}
    
\end{table}

\begin{table*}[t]

    \centering
    \resizebox{1.0\linewidth}{!}{
    
    \begin{tabular}{cl|c|cccccc|cc}
    \toprule
    
    & \multicolumn{1}{c}{} & \multicolumn{2}{c}{\textbf{In-domain}} 
    & \multicolumn{5}{c}{\textbf{Out-of-domain}} & \multicolumn{2}{c}{\textbf{Avgs}} \\ 
    \cmidrule(lr){3-4} \cmidrule(lr){5-9} \cmidrule(lr){10-11} 
    
    & Model & AIDA$_{dev}$ & AIDA$_{test}$ & MSNBC & AQUAINT & ACE2004 & CWEB & WIKI & Avg & Avg$_{OOD}$   \\
    
    \midrule
    \midrule
    
    \parbox[t]{2mm}{\multirow{3}{*}{\rotatebox[origin=c]{90}{\textit{AIDA+}}}}
    
    & \citet{yang-etal-2018-collective}     & - & \textbf{95.9}   & 92.6 & 89.9 & 88.5 & \textbf{81.8}   & 79.2    & 88.0    & 86.4  \\
    & GENRE                                 & - & 93.3   & 94.3 & 89.9 & 90.1 & 77.3   & 87.4 & 88.8 & 87.8 \\
    & ExtEnD$_{large}$                        & - & 92.6   & \textbf{94.7} & \textbf{91.6} & \textbf{91.8} & 77.7   & \textbf{88.8} & \textbf{89.5} & \textbf{88.9} \\
    
    \midrule
    \midrule
    
    \parbox[t]{2mm}{\multirow{7}{*}{\rotatebox[origin=c]{90}{\textit{AIDA}}}}
    
    & GENRE                                 & - & 88.6   & 88.1 & 77.1 & 82.3 & 71.9   & 71.7 & 79.5 & 78.2 \\
    & ExtEnD$_{base}$                         & - & 87.9   & 92.6 & 84.5 & \textbf{89.8} & 74.8  & 74.9 & 84.1  & 83.3 \\
    & ExtEnD$_{large}$                        & - & 90.0	& \textbf{94.5} & \textbf{87.9} & 88.9 & \textbf{76.6}	 & 76.7	& \textbf{85.8} & \textbf{84.9} \\
    
    \cmidrule(lr){2-11}
    
    & GENRE$^{\dagger}$                                 & 94.8 & 90.7 & 91.3 & 76.9 & 87.3 & 73.9 & 73.7 & 82.3 & 80.6 \\
    & GENRE$^{def}$                           & 93.2 & 84.4 & 83.1 & 59.6 & 81.3 & 64.0 & 63.4 & 72.6 & 70.3 \\
    & ExtEnD$_{base}^{def}$                   & 93.9 & 89.1 & 93.5 & 84.9 & 87.7 & 74.9 & 74.5 & 84.1 & 83.1 \\
    & ExtEnD$_{large}^{def}$                  & \underline{\textbf{94.9}} & \textbf{92.4} & 93.2 & 87.0 & 87.7 & 76.4 & \underline{\textbf{78.3}} & \textbf{85.8} & 84.5 \\
    
    \bottomrule
    \end{tabular}
        
    }
    
    \caption{\textit{inKB Micro $F_1$} scores over the AIDA-CoNLL validation and test splits, and the out-of-domain datasets when training on AIDA-CoNLL (bottom) or additional resources as well (top). The best score in each section is marked in \textbf{bold} and, in the bottom part, if its difference to its best alternative is statistically significant ($p < 0.01$ according to the McNemar's test \cite{dietterich1998approximate}), we also \underline{underline} it.}
    \label{tab:ed-results}
    
\end{table*}

In order to assess the applicability of our proposal to ED, we evaluate how the performances of generative and extractive formulations change when moving from Wikipedia titles to our alternative. To this end, in this Section, we first describe our experimental setting, discussing the datasets, evaluation strategy and comparison systems we adopt. Then, we describe the architecture we use for the two formulations. Finally, we present our findings.

\subsection{Experimental Setup}

\paragraph{Data} We follow the same experimental setting depicted by \citet{decao2020autoregressive} and use the standard AIDA-CoNLL splits \cite[AIDA]{hoffart-etal-2011-robust} for training, model selection and in-domain evaluation;
similarly, we leverage their cleaned version of MSNBC, AQUAINT, ACE2004, WNED-CWEB (CWEB) and WNED-WIKI (WIKI) \cite{guo2018robust,gabrilovich2013Freebase} for out-of-domain evaluation and use their same candidate sets, which were originally presented by \citet{le-titov-2018-improving}.\footnote{These candidate sets were generated through count statistics from Wikipedia, YAGO and a large Web corpus.}
We retrieve the description of each entity candidate through Wikidata\footnote{We took the latest dump (June 13th, 2022) at the moment of writing from the official Wikidata website: \href{https://dumps.wikimedia.org/wikidatawiki/entities/}{https://dumps.wikimedia.org/wikidatawiki/entities/}} and report in Table~\ref{tab:datasets} the number of instances and candidates in each dataset under consideration.
Due to inconsistencies in the datasets and different dump versions, the mapping from title to description is not always possible and, in these cases, we fall back to employing their Wikipedia title alone. 

\paragraph{Evaluation} Following previous literature in ED, we report scores over the test sets in terms of \textit{inKB Micro $F_1$}. Furthermore, for each system we consider, we report the average of its performances both over all the test sets ($\text{Avg}$) and over the five out-of-domain datasets only ($\text{Avg}_{\text{OOD}}$).

\paragraph{Comparison Systems} We consider the original models presented by \citet[GENRE]{decao2020autoregressive} and \citet[ExtEnD]{barba-etal-2022-extend}, trained on AIDA-CoNLL with Wikipedia titles, as our main natural comparison systems; in particular, for ExtEnD, we evaluate against both its Longformer base ($\text{ExtEnD}_{\text{base}}$) and large ($\text{ExtEnD}_{\text{large}}$) alternatives. Furthermore, to better contextualize the performances we attain within the current landscape of ED, we also include three state-of-the-art systems, namely, the global model of \citet{yang-etal-2018-collective} and the variants of \citet{decao2020autoregressive} and \citet{barba-etal-2022-extend} that were pre-trained on BLINK \cite{wu-etal-2020-scalable} before fine-tuning on AIDA-CoNLL. However, we note that, differently from our work, these three systems used additional training data (\num{9}M samples) from Wikipedia, whereas, due to computational constraints, we limit our analysis to the sole usage of AIDA-CoNLL ($<20$K samples).

\subsection{Architectures}

For both our formulations, we closely follow the corresponding reference architectures. For the generative methods, we use BART (\num{406}M parameters) as our underlying sequence-to-sequence model and fine-tune it on AIDA-CoNLL. As for the extractive approach, we test and evaluate our approach on both the \textit{base} (\num{139}M parameters) and \textit{large} (\num{435}M parameters) versions presented in the reference work. We report training details in Appendix \ref{sec:appendix-training}.

\subsection{Results}

\begin{table}[t]

    \centering
    \resizebox{1.0\linewidth}{!}{
    
        \begin{tabular}{clrrrrr}
        \toprule
        & Model & MFC & LFC & UE & UEM & UM \\
        
        \toprule
        
        \parbox[t]{1mm}{\multirow{2}{*}{\rotatebox[origin=c]{90}{\textit{AIDA}}}}
        & ExtEnD$_{large}$            & \textbf{98.3} & \textbf{81.6} & 80.9 & 80.9 & 89.0 \\
        & ExtEnD$_{large}^{def}$      & \textbf{98.3} & 81.0 & \underline{\textbf{86.9}} & \underline{\textbf{86.5}} & \underline{\textbf{92.9}} \\
        
        \midrule
        
        \parbox[t]{1mm}{\multirow{2}{*}{\rotatebox[origin=c]{90}{\textit{OOD}}}}
        & ExtEnD$_{large}$            & \textbf{97.2} & 82.2 & 73.8 & 74.4 & 77.2 \\
        & ExtEnD$_{large}^{def}$      & 96.5 & 81.5 & \underline{\textbf{74.5}} & \textbf{75.0} & \textbf{77.7} \\
        
        \bottomrule
        \end{tabular}
        
    }
    
    \caption{Fine-grained results analysis over the AIDA-CoNLL (top) and out-of-domain (bottom) datasets. 
    }
    \label{tab:fine-grained}
    
\end{table}

In Table~\ref{tab:ed-results} we show the \textit{inKB Micro $F_1$} score that our models and its comparison systems achieve on the datasets under consideration. As a first note, we point out that, for easier comparability in our experiments, we reproduced the original AIDA-CoNLL models of both \citet{decao2020autoregressive} and \citet{barba-etal-2022-extend}. While we attain comparable performances for the latter, and hence omit it, we find that our $\text{GENRE}^{\dagger}$ implementation obtains better results than its reference, especially out of domain, with an average improvement of more than \num{2} points.

Moving to $\text{GENRE}^{def}$, its behavior is definitely below its counterpart with Wikipedia titles, with a drop of roughly \num{10} points on average. To better understand this issue, we analyzed its predictions over the validation set but did not identify any significant error pattern. In particular, we investigated whether $\text{GENRE}^{def}$ presented length biases or was excessively skewed towards the most frequent entities and, consequently, less apt to scale over least frequent entities or unseen mentions; interestingly, we did not find either of these to be the case, with the two systems having similar error distributions. We believe instead that the drop might be happening as the formulation behind $\text{GENRE}^{def}$ requires modeling a much more complex output space and more data could be needed to properly scale. However, besides this negative finding, $\text{GENRE}^{def}$ presents an additional issue that does not show through Table~\ref{tab:ed-results}. Indeed, while using Wikipedia titles results in output sequences with an average subword length over AIDA-CoNLL of \num{7} and {99}th percentile of \num{14}, adding descriptions results in considerably longer entity representations: the average nearly doubles, reaching \num{12.5}, while the \num{99}th percentile hits \num{29}. In turn, this implies longer prediction times, which might make this formulation unfeasible in some practical settings.

Considering instead extractive formulations, we find the role of definitions to be definitely more impactful. $\text{ExtEnD}_{base}^{def}$ surpasses $\text{ExtEnD}_{base}$ on \num{3} out of {5} out-of-domain benchmarks and on the standard test set, here by more than \num{1} point. Besides, while the two systems achieve comparable $\text{Avg}$ and $\text{Avg}_{\text{OOD}}$ scores, this is mostly due to the ``large'' drop in ACE2004, which counts less than \num{260} instances but still negatively affects $\text{ExtEnD}_{base}^{def}$ macro behavior. However, arguably our most interesting finding is the behavior of $\text{ExtEnD}_{large}^{def}$, which  
attains statistically significant improvements on AIDA-CoNLL ($+2.4$) and WIKI ($+1.5$), and comparable performances on CWEB; note that these three datasets are, by far, the largest benchmarks in our experimental setup (Table~\ref{tab:datasets}). 

Furthermore, we investigate the effectiveness of $\text{ExtEnD}_{large}^{def}$ over different classes of label frequency, both in-domain (AIDA-CoNLL) and out-of-domain (concatenation of the five datasets), and compare it with $\text{ExtEnD}_{large}$ (Table~\ref{tab:fine-grained}). Specifically, we consider instances
\begin{enumerate*}[label=\roman*)]
    \item tagged with their most frequent entity (MFC) in the training set,
    \item tagged with a least frequent entity (LFC),
    \item tagged with an unseen entity (UE),
    \item whose (mention, entity) pair (UEM) or
    \item whose mention (UM) does not appear in the training set.
\end{enumerate*}
Overall, apart from the MFC and LFC classes, where the difference is not statistically significant, $\text{ExtEnD}_{large}^{def}$ fares better in all other settings, which all require scaling over unseen patterns. Most notably, it yields $+6.0$ (AIDA) and $+0.7$ (OOD) improvements, both statistically significant, on unseen entities. This underlines the better generalization capability granted by the use of more expressive textual representations.

\section{Conclusion}

In this work, we focus on a shortcoming of generative and extractive formulations to Entity Disambiguation, namely their usage of Wikipedia titles, which are often insufficiently informative, and explore the effect of more expressive representations on these formulations. While we do not witness positive gains for generative formulations, at least in the limited data and computational regime we consider, we report strong improvements on extractive formulations. Specifically, our extractive approach sets a new state of the art on \num{2} out of the \num{6} benchmarks under consideration and, more interestingly, shows better scalability over unseen patterns, especially unseen entities.

\section*{Limitations}

We believe that our work has three major limitations. First, both the generative and extractive formulations that we consider lack parallelism, as they disambiguate each mention in the input text one at a time. While batching can definitely help, it poses additional computational requirements and, besides, the same (but for the position of the \textit{<s>} and \textit{</s>} special symbols) input text would still need to be encoded multiple times. Second, our representation strategy requires the availability of descriptions in the target language in Wikidata (or some other knowledge base with a mapping from Wikipedia titles). While this data was readily available for English, this might not be the case for several other mid-to-low-resource languages. Finally, both our formulations are local and, granted that pre-trained language models have certainly bridged the gap with global alternatives, their underlying independence assumption is still limiting.

\section*{Acknowledgments}
\begin{center}
\noindent
    \begin{minipage}{0.1\linewidth}
        \begin{center}
            \includegraphics[scale=0.2]{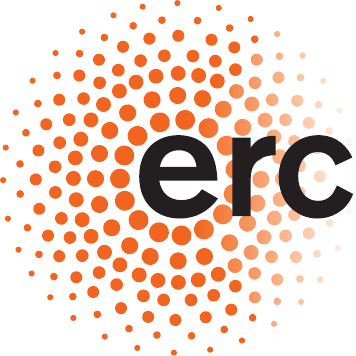}
        \end{center}
    \end{minipage}
    \hspace{0.01\linewidth}
    \begin{minipage}{0.70\linewidth}
        The authors gratefully acknowledge the support of the ERC Consolidator Grant MOUSSE No.\ 726487 under the European Union's Horizon 2020 research and innovation programme.
    \end{minipage}
    \hspace{0.01\linewidth}
    \begin{minipage}{0.1\linewidth}
        \begin{center}
            \includegraphics[scale=0.08]{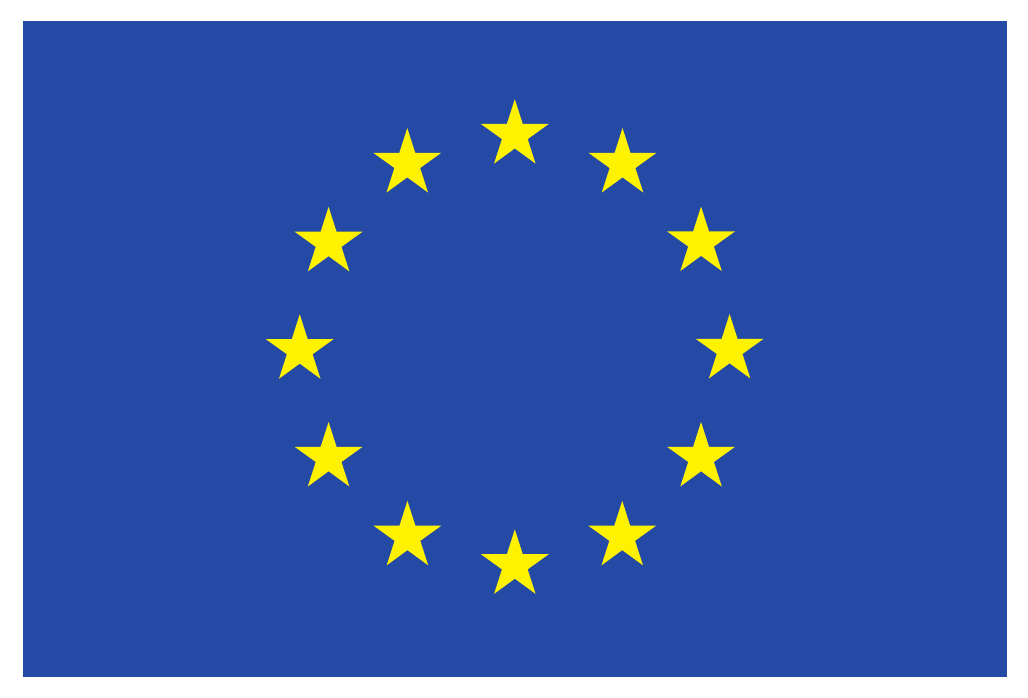}
        \end{center}
    \end{minipage}\\
\end{center}

\bibliography{anthology,custom}
\bibliographystyle{acl_natbib}

\appendix

\section{Training Details}
\label{sec:appendix-training}

For both our formulations, we closely follow the original works in terms of training procedures. In particular, for the generative methods, we fine-tune BART (\num{406}M parameters) on AIDA-CoNLL using \num{10000} effective token batch size, Adam \cite{DBLP:journals/corr/KingmaB14} as our optimizer and $10^{-5}$ learning rate, with $500$ warm-up steps and linear decay. As for the extractive approach, we include both the \textit{base} (\num{139}M parameters) and \textit{large} (\num{435}M parameters) versions presented in the reference work, use Rectified Adam as our optimizer, with $10^{-5}$ learning rate, and train with an effective token batch size of \num{8000} tokens. All the trainings are done for a single run on GeForce RTX \num{3090} graphic card with \num{24} gigabytes of VRAM.


\end{document}